 \documentclass[pmlr,twocolumn,10pt]{jmlr} 

\usepackage{booktabs}
\usepackage{siunitx}

\theorembodyfont{\upshape}
\theoremheaderfont{\scshape}
\theorempostheader{:}
\theoremsep{\newline}

\jmlrvolume{LEAVE UNSET}
\jmlryear{2023}
\jmlrsubmitted{LEAVE UNSET}
\jmlrpublished{LEAVE UNSET}
\jmlrworkshop{Machine Learning for Health (ML4H) 2023} 

\title[De-Identification of clinical notes using Spark NLP]{Beyond Accuracy: Automated De-Identification of Large Real-World Clinical Text Datasets}
 
\author{
\Name{Veysel Kocaman} \Email{veysel@johnsnowlabs.com}\\
\Name{Hasham Ul Haq} \Email{hasham@johnsnowlabs.com}\\
\Name{David Talby} \Email{david@johnsnowlabs.com}\\
\addr John Snow Labs Inc., Delaware, USA
 }

\begin{document}

\maketitle

\begin{abstract}
Recent research advances achieve human-level accuracy for de-identifying free-text clinical notes on research datasets, but gaps remain in reproducing this in large real-world settings. This paper summarizes lessons learned from building a system used to de-identify over one billion real clinical notes, in a fully automated way, that was independently certified by multiple organizations for production use.

A fully automated solution requires a very high level of accuracy that does not require manual review. A hybrid context-based model architecture is described, which outperforms a Named Entity Recogniton (NER)-only model by 10\% on the i2b2-2014 benchmark. The proposed system makes 50\%, 475\%, and 575\% fewer errors than the comparable AWS, Azure, and GCP services respectively while also outperforming ChatGPT by 33\%. It exceeds 98\% coverage of sensitive data across 7 European languages, without a need for fine tuning.

A second set of described models enable data obfuscation – replacing sensitive data with random surrogates – while retaining name, date, gender, clinical, and format consistency. Both the practical need and the solution architecture that provides for reliable \& linked anonymized documents are described.
\end{abstract}

\section{Introduction}
\label{sec:intro}

Electronic Health Records (EHRs) are now in use by more than 96\% of acute care hospitals and 86\% of office-based physicians in the USA \cite{myrick2019percentage}. While most billing \& claims data is in structured format, a lot of clinical data - e.g. progress notes, discharge summaries, radiology reports, pathology reports - are in the form of unstructured text. Making this data available to research bodies is vital for a variety of secondary uses like population health, real-world evidence, patient safety, and drug discovery. 

Since this data can contain highly sensitive information, it must first go through a de-identification process. De-identified patient data is defined as health information that has been stripped of all “direct identifiers” — that is, any information that can be used to uniquely identify the patient. The Health Insurance Portability and Accountability Act (HIPAA)'s Safe Harbour guidelines define 18 such direct identifiers \cite{us2003hipaarule} \cite{stubbs2017identification} - although any other data point that can uniquely identify a patient must be considered as well.

After defining how a specific dataset should be de-identified, the task of identifying protected health information (PHI) in structured or unstructured data can be automated. A recent meta-review by \cite{Yogarajan2020ARO} found 18 papers of systems that achieve an F-measure above 95\% on the 2014 i2b2 de-identification challenge \cite{stubbs2015automated}. This threshold is widely regarded as matching the accuracy of manual de-identification \cite{stubbs2015automated}. \cite{Neamatullah2008AutomatedDO} reported that a single human annotator achieved a recall of 81\% while requiring a consensus of two human annotators raised it to 94\%.

However, the same meta-review also highlighted gaps in the proposed systems - concluding that a high F1-measure is a necessary but insufficient condition for enabling automated de-identification on real-world clinical text. Two areas of concern were common consistent mistakes that systems make, and challenges in obfuscating PHI in a medically consistent manner.

This paper presents lessons learned from a system that has addressed this and other practical challenges on real-world clinical documents. This system has been used over the past five years in production settings, has been used to automatically de-identify over one billion real-world clinical notes \cite{Talyab2022Lessons}, and has been certified through an independent expert determination process by multiple organizations and in multiple countries \cite{Piotrowski2022}. Handling this scale of data volume and variety required automated solutions to longstanding pragmatic challenges. Specifically, the contributions of this paper are:

\begin{itemize}
  \setlength{\itemsep}{0.5pt}
    \item Introduce a natively scalable, pre-trained NLP pipeline that delivers state-of-the-art accuracy on academic benchmarks and outperforms the accuracy of the commercial services of the 3 major public cloud providers as well as ChatGPT \cite{ChatGPT}.
    \item Specify how to achieve the same level of accuracy in multiple languages with minimal effort, beyond the 7 languages the system currently supports.
    \item  Propose a method for data obfuscation - the task of replacing PHI with medically appropriate random surrogates - filling six data consistency requirements.
\end{itemize}

\section{Background and Related Work}

De-identification of unstructured data is a well-studied problem, and various Natural Language Processing (NLP) \cite{nadkarni2011natural} approaches have been proposed till date \cite{khin2018deep}. It can be split into two subtasks: First, PHI needs to be identified in text, and second, those identifiers are then replaced using masking (with a placeholder value) or obfuscation (with a random value based on its type). The first subtask has received more attention in research. 

Early de-identification systems in the clinical domain were mainly rule-based, such as \cite{sweeney1996replacing} and \cite{gupta2004evaluation}. These systems employed regular expressions, syntactic rules, and specialized dictionaries to identify PHI in text. Rule-based systems usually perform well on recognizing formulaic PHI instances - i.e. phone numbers, emails, licenses, etc. - but struggle with concepts like the names of people, professions, and hospitals \cite{liu2017identification}. Rule-based systems also require major changes in dictionaries and rules when implemented in new environments, generalising poorly over unseen datasets.

The concept of automatic de-identification was first introduced into the Informatics for Integrating Biology and the Bedside (i2b2) project \cite{uzuner2007evaluating} in 2014, an academic NLP challenge on automatically detecting PHI identifiers from medical records. These challenges have boosted research and development of Machine \& Deep Learning algorithms for robust PHI identification.  Conditional Random Fields (CRF) \cite{he2015crfs} and hand-crafted features using lexical rules \cite{lafferty2001conditional} to identify required concepts from data were among the most popular early approaches.

As vector based language models became more efficient in encapsulating semantic information, the trend shifted to deep learning (DL) models leveraging semantic information from language models. \cite{liu2017identification} used a hybrid system comprising of CRF, Bi-LSTM, Word2Vec, and dictionaries for Named Entity Recognition (NER) on clinical notes. While Bi-LSTM and embedding-based models have better generalisation ability, they too have limitations when it comes to extracting large chunks \cite{yang2019study}. 

To address this, further studies have included more syntactic and semantic information while being cognizant of temporal information. \cite{Yogarajan2020ARO} reports that of 18 published systems that achieve an F-measure of 95\% or more of academic benchmarks, 10 were hybrids of rules and models, and 8 used only models.

\section{Architecture}
\label{sec:implementationec}

\subsection{Scalable NLP Pipeline}

The proposed system is built on top of the Spark NLP library \cite{kocaman2021spark}, a popular open-source library that supports most NLP tasks and can uniquely scale up both training and inference on any Apache Spark cluster. 
The system can be deployed on a single machine (local mode) or on an Apache Spark cluster with no code changes. De-identification is implemented as an NLP pipeline - a sequence of processing steps - that runs as one end-to-end solution including text pre-processing, deep-learning based models, contextual rules, and obfuscation. The pipeline was externally benchmarked to de-identify 500,000 patient notes in 2.46 hours on 10 single-CPU commodity servers \cite{tomer2021}. No code changes are required to scale to a cluster \cite{kocaman2021spark}.

The pre-trained de-identification pipelines can be broken down into 5 stages: text pre-processing and feature generation, named entity recognition, contextual rules, chunk merging, and obfuscation. Strong data to enable future re-identification is also included.

\begin{figure}[h]
  \centering
  \includegraphics[width=\linewidth]{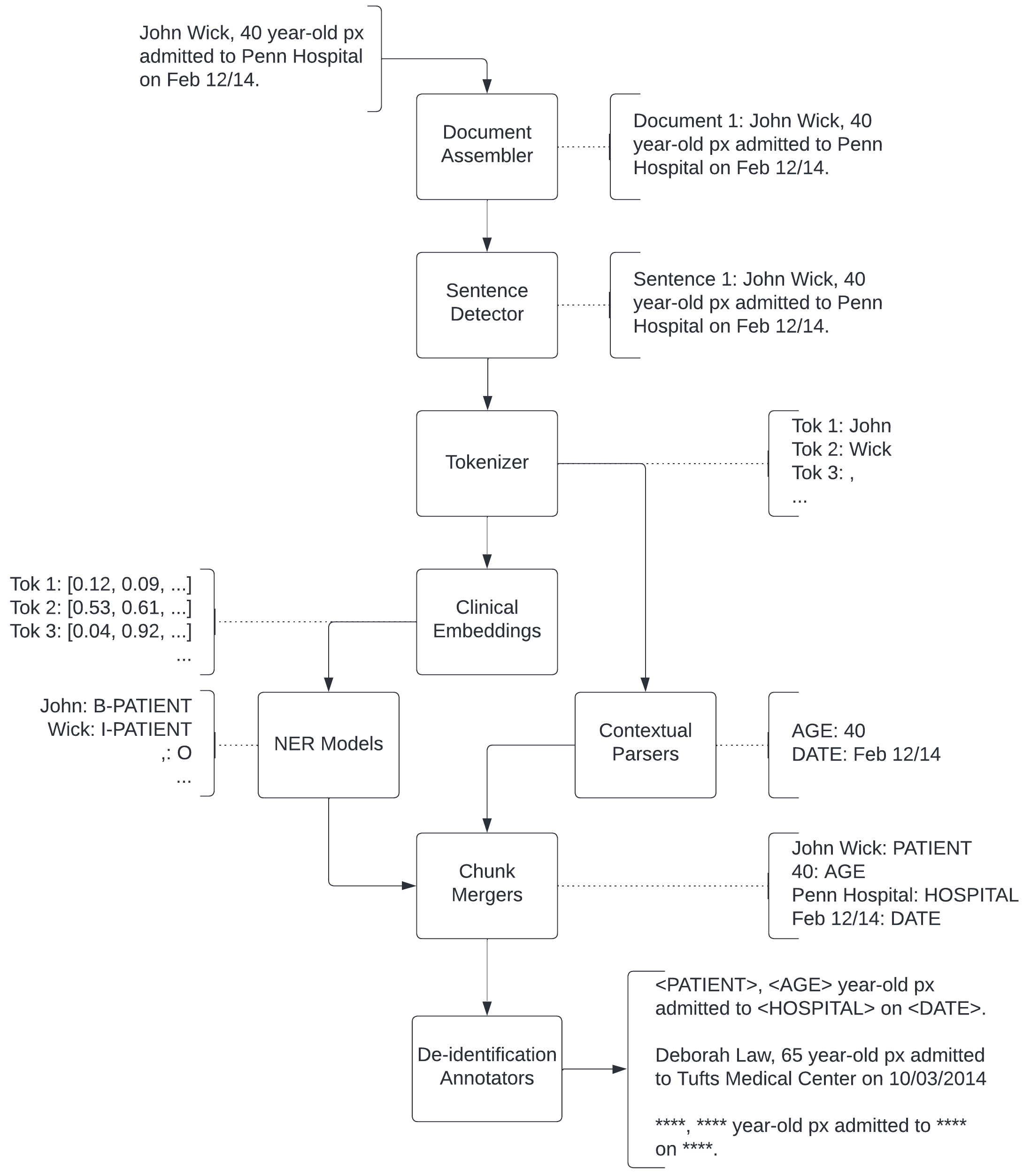}
  \caption{Full pipeline architecture}
  \label{fig:architecture}
\end{figure}

\subsubsection{Text Pre-Processing}

The input text must first pass through a series of stages which include a document assembler, sentence detector, tokenizer, and word embeddings generator. These pipeline components create the appropriate features for subsequent components to identify PHI tokens and then obfuscate PHI chunks.

The first step is the document assembler that takes in raw text data and creates the first annotation which can then be used as inputs to downstream tasks. The next step in the pipeline involves sentence detection. This is based on a general-purpose  deep-learning based model \cite{schweter2019deep} for sentence boundary detection. The model was tuned specifically for clinical text in English, and multilingual text for other languages. Using rule-based sentence boundary detection performs poorly on real-world clinical notes, which often do not use standard punctuation and grammar.

The next step is tokenization, which involves splitting sentences into smaller sub-units to generate meaningful features for the NER task. The tokenization method used in de-identification pipelines is a rule-based tokenization which separates characters on white spaces and other special characters. Once the input text has been broken down into tokens, each token gets assigned a word embedding feature vector. The word embeddings used in the de-identification pipelines are custom-trained using a skip-gram model on PubMed abstracts and case studies for learning distributed representations of words using contextual information. The trained word embeddings have a dimension of 200 and a vocabulary size of 2.2 million.

\subsubsection{Named Entity Recognition (NER)}

The NER model is the next step in the pipeline. Its goal is to detect PHI elements such as names of patients, doctors, organizations, hospitals, streets, cities, countries, professions, dates, ages, and others. NER models are at the core of de-identification pipelines, as they have can generalize on unseen data better, and predict exact entity spans of PHI elements, causing minimum data loss. Our NER model implementation is based on a BLSTM architecture as explained in \cite{kocaman2021biomedical}.

Using a combination of publicly available resources and in-house datasets, we curated two versions of NER training datasets and then trained two NER models for each of the seven supported languages. These include a coarse version with seven entity types \textit{(Name, Date, Organization, Location, Age, Contact, ID)} detected and a granular version with thirteen \textit{(Patient, Doctor, Hospital, Date, Age, Profession, Organization, Street, City, Country, Phone, Username, Zip)} entity types. 

\subsubsection{Contextual Rule Engine}

While NER models generalize better than rule engines, there are cases where rule engines can provide much needed flexibility. For example, different geographical, political, administrative regions, and organizations have different unique identifier numbers (e.g. Patient IDs, License Number, Phone Numbers etc). Including a regex based parser in the same NLP pipeline provides the ability to complement NER models in cases where a specific PHI identifier type is not supported by the NER model, without retraining the NER models.

In addition to regex matching, the contextual rule engine supports prefix and suffix matching for reducing false positives, and the proximity and length of context matching can also be adjusted to avoid false positives. For example, \textit{age} can be a numeric number, but the unit of age (e.g month, year, etc) would be suffix matching within a short span within the context string.

Another benefit of having the rule engine and NER model in the same pipeline is to reduce label errors by the NER model. For example, the NER model can get confused between patient ID numbers and medical record numbers, but the contextual rule engine using prefix and regex matching of the contextual rule engine's prefix and suffix matching capability can be used to develop heuristics, depending on document layout.

\subsubsection{Chunk Merger}

Up to this point in the pipeline, PHI chunks will have been identified by both models and rules. The next step is to merge these outputs together to resolve conflicting or overlapping detections and increase the overall accuracy of the pipeline. 

The chunk merger does this by assigning a priority to each entity type detected by each model type. This is configurable based on the needs of each use-case. For example, a SSN identified by rules would usually get higher priority over any overlapping chunk identified by the NER model. 

\subsubsection{Masking or Obfuscation}

The final step of the pipeline is called de-identification and its role is to generate the anonymized text. This can be achieved by applying either masking or obfuscation. Masking essentially replaces the PHI identifier with either their type or asterisks. These asterisks can either be of fixed character length for all PHI chunks detected in the text, or of the same length as the PHI chunk being replaced (see Figure \ref{fig:architecture}); we found the later option to be helpful while de-identifying pdf and image documents, as it minimizes any changes to the original document layout. 

Obfuscation involves replacing PHI with surrogate values that are semantically, and linguistically correct. The following example shows the original (identifiable) text followed by the same text masked by entity type, masked by asterisks, masked with the age field white-listed, and finally obfuscated:
\\

 \textit{Jane is a 48-year-old nurse from Memphis.}
 
 \textit{PATIENT is a AGE PROFESSION from CITY.}
 
 \textit{*** is a *** *** from ***.}
 
 \textit{*** is a 48-year-old *** from ***.}
 
 \textit{Gina is a 45-year-old teacher from Fresno.}\\

Since the system is designed to support de-identification of PDF, DOCX, DICOM, and Image files, preserving the text layout is important. Obfuscating the text with surrogate values having similar length is relatively challenging that simple masking, therefore, we divided the vocabularies in multiple groups based on character length for faster search during inference. The module first searches for the matching group - based on chunk length - and then applies logic additional logic to select best surrogate values that maintain data integrity as explained in section \ref{sec:consistentobf}.

\subsubsection{Re-Identification Vault}

Should users require the ability to re-identify de-identified PHI text - for example, to support emergency unblinding - the pipeline provides an option to save the mappings between obfuscated/masked chunks and their original form in an auxiliary column. The mapping can then be saved as a Parquet file (a column-oriented data storage format), which can later be used as input to a re-identification pipeline. The Re-Identification pipeline uses the saved mapping information to regenerate the original text.

\section{Experimental Setup \& Results}

This section describes benchmark datasets, evaluation metrics, and an overview of the setup. We conducted two experiments:

\begin{enumerate}
  \item Accuracy evaluation on the 2014 i2b2 de-identification challenge dataset \cite{stubbs2015automated}.
  \item Using a sample from MIMIC-III dataset \cite{johnson2016mimic}, benchmarking against commercial cloud API's.
\end{enumerate}

\subsection{Datasets}

Datasets containing PHI data are a prerequisite for training and evaluating de-identification solutions. However, for privacy reasons, they are rarely released to the public. A few exceptions exist, where records that have been de-identified are released by organizations or researchers to advance the field of clinical de-identification. 

Our first experiment made use of a dataset released with the 2014 i2b2/UTHealth shared task Track 1 de-identification challenge \cite{stubbs2015automated} which tests the performance of automated systems on PHI concepts. Our second experiment consisted of an out-of-the-box comparison of our de-identification solution against commercially available de-identification solutions.

To conduct a fair comparison, and without knowledge of the data sources used to train these licensed models, we can confidently assume that the data used to train these systems was a mix between proprietary in-house annotated data and publicly available datasets. Keeping that assumption in mind, and to ensure reproducibility, we randomly sampled one hundred clinical notes as a test set from the same i2b2 de-identification dataset mentioned above and tasked two medical doctors to annotate them and populate ground truths for 8 types of entities: Date, Age, Doctor, Patient, Hospital, ID, Location and Phone. These categories were chosen to accommodate the differences in extraction categories between the different de-identification solution providers. 

Experiments were run in a Google Colab server (2vCPU @ 2.2 GHz, 13 GB RAM), using Apache Spark in local mode. 

\subsection{Accuracy on the 2014 i2b2 De-Identification Challenge}

In the first experiment, using the original training set, we trained two NER models (coarse 7 labels, granular 13 labels) with no additional rules and component (e.g. regex matchers, contextual parsers) and evaluated on the official test set and achieved 0.955 and 0.978 micro F1 scores respectively. The detailed metrics for 7-labels version can be seen at Table \ref{tab:i2b2_stats} and the metrics for granular 13-labels version can be seen in \appendixref{apd:first}.

\begin{table}[hbt!]
\caption{NER metrics on 2014 i2b2 Challenge using 7 labels.}
\label{tab:i2b2_stats}
\scalebox{0.95}{
\begin{tabular}{lccc}
\toprule
Entity & Precision & Recall & F1\\
\midrule
Contact &  0.958 & 0.961 & 0.959\\
Name & 0.968 & 0.956 & 0.962\\
Date & 0.991 & 0.980 & 0.985\\
ID & 0.964 & 0.895 & 0.928\\
Location & 0.941 & 0.902 & 0.921\\
Profession & 0.889 & 0.601 & 0.719\\
Age & 0.955 & 0.936 &  0.945\\
\midrule
\textbf{Macro-Avg.} & & & \textbf{0.917}\\
\textbf{Micro-Avg.} & & & \textbf{0.955}\\
\bottomrule
\end{tabular}}
\end{table}

\subsection{Comparison with Commercial Cloud API's}

In the second experiment, we evaluated three of the most widely known commercial services for de-identification that provide APIs through which users can send sensitive records to identify PHI: AWS Medical Comprehend (AMC), Google Cloud Platform Healthcare API (GCP), and Microsoft Azure Text Analytics for Health (Azure).  

After standardizing the tags to accommodate the different commercial APIs, the system ran on the one hundred clinical notes from the i2b2 corpus. Our solution and AMC had good coverage over the tags, whereas Azure and GCP required using a combination of their de-identification APIs in combination with other NLP APIs for covering entities such as locations and names. 

To accommodate the potential differences in chunk boundaries identified by the different systems, we set a threshold of 60\% where each system was given a valid detection if it covered at least 60\% of the annotated chunk. For example, detecting \textit{"Children's Hospital"} versus the ground truth being \textit{"Boston Children's Hospital"} was assumed correct. This experiment is fully reproducible and will be made available online. Azure and GCP did not provide the ability to identify ID numbers appropriately, so ID entities were excluded from the metric calculations. 

Results show that our de-identification pipeline outperformed all commercial APIs (see Table\ref{tab:bench_full}). Our system yielded an average F1 score of 0.96, whereas AWS Medical Comprehend and Microsoft Azure Text Analytics for Health and Google Cloud Platform Healthcare API obtained average F1 scores of 0.94, 0.81 and 0.77 respectively. 

\begin{table*}
\caption{Comparison of the de-identification pipeline with AWS Medical Comprehend (AMC), Microsoft Azure Text Analytics for Health (Azure), \& Google Cloud Platform (GCP) Healthcare API on a sample of 100 notes.}
\scalebox{0.75}{
\label{tab:bench_full}
\renewcommand{\arraystretch}{1.3}
\begin{tabular}{llcccccccccccc}
\toprule
\multicolumn{2}{l}{} & \multicolumn{3}{c}{\textbf{Ours}} & \multicolumn{3}{c}{\textbf{AMC}} & \multicolumn{3}{c}{\textbf{Azure}} & \multicolumn{3}{c}{\textbf{GCP}} \\ \cline{3-14}
Entity & Sample & Precision & Recall & F1 & Precision & Recall & F1 & Precision & Recall & F1 & Precision & Recall & F1 \\ \midrule 
Age & 95 & 1.000 & 1.000 & 1.000 & 0.989 & 0.936 & 0.962 & 0.882 & 0.976 & 0.927 & 0.888 & 0.929 & 0.908 \\
Date & 953 & 0.999 & 0.995 & 0.997 & 1.000 & 0.990 & 0.995 & 1.000 & 0.811 & 0.896 & 1.000 & 0.928 & 0.962 \\
Doctor & 402 & 0.987 & 0.969 & 0.978 & 1.000 & 0.918 & 0.957 & 0.987 & 0.551 & 0.707 & 0.503 & 0.749 & 0.602 \\
Hospital & 182 & 0.922 & 0.911 & 0.917 & 0.980 & 0.810 & 0.887 & 0.962 & 0.573 & 0.718 & 0.634 & 0.829 & 0.718 \\
Location & 47 & 0.905 & 0.884 & 0.894 & 0.842 & 0.780 & 0.810 & 1.000 & 0.766 & 0.867 & 0.614 & 0.900 & 0.730 \\
Patient & 115 & 1.000 & 0.930 & 0.964 & 1.000 & 0.904 & 0.950 & 0.949 & 0.667 & 0.783 & 0.545 & 0.424 & 0.477 \\
Phone & 15 & 1.000 & 1.000 & 1.000 & 1.000 & 1.000 & 1.000 & 1.000 & 0.667 & 0.800 & 1.000 & 0.933 & 0.966 \\
ID & 465 & 0.941 & 0.910 & 0.925 & 0.922 & 0.933 & 0.927 & -     & -     & -     & -     & -     & -    \\ 
\midrule
\multicolumn{2}{l}{\textbf{Macro-Avg.}} & & & \textbf{0.959} & & &  \textbf{0.936} & & & \textbf{0.712} & & & \textbf{0.670}\\
\multicolumn{2}{l}{\textbf{Micro-Avg.}} & & & \textbf{0.969} & & &  \textbf{0.959} & & & \textbf{0.715} & & & \textbf{0.638}\\
\bottomrule
\end{tabular}}
\end{table*}

\subsection{Comparison with ChatGPT (GPT3.5)}

On a selected 25 clinical notes from i2b2 dataset, our approach demonstrates superior performance with a 93\% accuracy rate compared to ChatGPT’s 60\% accuracy. To maintain the conciseness of this paper, additional details are presented in the  \appendixref{apd:third}.

\section{Key Modules}

Achieving high accuracy was not enough to deliver fully automated de-identification on large real-world datasets. This section describes three key areas in which innovation was required as this system should be capable of handling close to a billion pieces of clinical text around the world.

\subsection{Multilingual Support}

Sources of labeled training data for de-identification are limited in availability for English - and even more so for other languages. Training language-specific models for each language is still necessary, since current multi-lingual word embeddings do not provide high accuracy and are not tuned to each country's medical vocabularies.

The proposed de-identification system is designed to be extendable to other languages with minimal effort. Adding a language typically requires 1-2 weeks, with the process being as follows:

\textbf{NER:} Achieving coverage over seven languages required re-purposing data commonly used for other tasks, and assembling different sources to cover all PHI entity types. Most of the PHI entity types can be found in the annotated datasets shared publicly by the community, such as ConLL \cite{sang2003introduction} and OntoNotes \cite{weischedel2011ontonotes}. These datasets include a subset of PHI entity types - Name, Location, Date - in many languages. However, these annotated datasets do not represent the contextual structure that a clinical note may have.
  
\textbf{Translation:} When data sources were exhausted, translation became another solutions to increase coverage and robustness of the training datasets. Whenever there is a coverage issue with certain entity type, we start with an annotated dataset from another similar language (i.e. English), mask the PHI entities in it (using the English model), then translate the entire sentence using Marian neural machine translator models \cite{junczys2018marian}, and then replaced the mask with the equivalence from other language (e.g. English names replaced by German names). This results with a brand new sentence having that entity (requiring a review by a German-speaking medical doctor).
  
\textbf{Contextual Rules:} 
The type and format of the entities differ from one country to another - even for countries that speak the same language, like the US and UK. This is because clinical documentation is taught, written, and used differently in each country. As a result, all the rules and patterns are manually modified, by a medical doctor from each country, to reflect each country's corresponding language.

Following these steps, and given that the NLP pipeline and each stage are modular and configurable, providing end-to-end support for a new language once its dataset is available requires only a few lines of code. Accuracy metrics for currently supported languages can be seen in \appendixref{apd:second}.

\subsection{Consistent Data Obfuscation}
\label{sec:consistentobf}

Obfuscation is more popular in practice than masking, for two reasons. First, it generates a more realistic looking result for testing, demos, or training downstream models. Second, it provides an extra layer of safety, since in contrast to masking, an attacker cannot tell when a PHI entity was missed. 

However, obfuscation comes with its own set of challenges, and is considered a barrier to real-world automated de-identification \cite{Yogarajan2020ARO}. For example, consider the following text: \\

\textit{Jane Doe is a lovely 78 y.o. lady with a history of breast cancer. Jane was diagnosed with T2DM in April 2020.} \\



Given this example, the obfuscation needs to get several things right to retain readability and consistency:

\begin{itemize}
  \setlength{\itemsep}{0.1pt}
  \item \textbf{Name consistency:} Mapping \textit{Jane Doe} and \textit{Jane} to the same first name. If we map \textit{Jane Doe} to a fake name (e.g. \textit{Nancy Smith}), the next name entity (\textit{Jane}) that corresponds to the same patient should also be replaced by \textit{Nancy}. In addition, when there are multiple clinical notes for the same patient, the same mapping should be made across different documents to have a consistent obfuscation for several concerns such as traceability or regulations. Hence, this mapping can be aligned to patient IDs so that every patient will get different mapping even for the same names (e.g. "Jane" will be mapped to "Mary" for patient-1 whereas the same name is mapped to "Jen" for patient-2). 
  \item \textbf{Gender consistency:} Mapping \textit{Jane} to a feminine American name (or a feminine British name if needed).
  \item \textbf{Age consistency:} Specifying a proper age range (i.e. age groups such as 5-12 years for children, 20-39 years for adults etc.) to make the obfuscation within that age group. The age obfuscation should be consistent here due to some phrases (e.g. \textit{lady}, \textit{lovely}) that hint to an adult lady. Hence we should replace \textit{78} with a reasonable age (e.g. \textit{40} but not \textit{5} or \textit{12}). 
  \item \textbf{Clinical consistency:} Note that \textit{Jane} needs to "remain female" also because she has a history of breast cancer.
  \item \textbf{Day shift consistency:} Shifting the days based on a predefined list of shift values per patient ID (e.g. plus 2 days for patient-1, minus 5 days for patient-2 etc.) as well as allowing a completely random shift given a range.
  \item \textbf{Date format consistency:} If \textit{April 2020} needs to be shifted by a random number of days, then the result should be in the same format (i.e. \textit{March 2020} and not \textit{3/3/2020}). Moreover, since there is no day information in the original date entity and it is not in a proper date format, this date should be normalized to a proper date format (e.g. \textit{04/1/2020}) at first in order to apply a day shift.
  \item \textbf{Length consistency:} In order to keep the length of the original text intact, it's often required to replace the selected entities with the same length of fake entities. If same length is not possible, adding or deleting characters can force it into the same length.
\end{itemize}

Two components were built to facilitate this. First is a normalization module that normalizes dates, ages, names, and addresses. It ensures that multiple occurrences of any concept are normalized to the same concept, even if they are mentioned differently throughout the text. In addition, date normalization is also enabling day shifting by normalizing the unstructured dates into proper format so that shift operation can be applied (e.g. 12Apr2022 will be normalized to 04/12/2024). 

Second is a \textit{faker} module that generates random data to replace original concepts. To maintain data integrity, the module is cognizant of people's titles, genders, and addresses to generate semantically correct values. As a second line of defence, we added the option of masking or total random obfuscation of concepts that can not be normalized, especially dates. Also, since date formats are particularly sensitive to geographical locations, we added the option of converting all dates to certain formats for consistency.

Like masking, the faker can make the text appear as close as possible to its original form (based on Levenshtein distance) to maintain formatting. This module also allows users to feed their own look-up dictionary for obfuscating the detected PHI chunks with custom replacements. Every field can be separated and configured by its own rules. The system comes with pre-built faker models for every language it supports, so that for example \textit{"Danilo Ramos de Madrid"} can be obfuscated to a \textit{"Jose Fernando de Alicante"} instead of to an American-biased name and city.

\begin{table*}
\caption{Comparison of NER models with full pipelines enriched with regex and contextual parser using Macro-F1 scores.}
\scalebox{0.70}{
\label{tab:nervspp}
\renewcommand{\arraystretch}{1.3}
\begin{tabular}{lcccccccccccccc}
\toprule
 & \multicolumn{2}{c}{\textbf{English}} & \multicolumn{2}{c}{\textbf{German}} & \multicolumn{2}{c}{\textbf{Spanish}} & \multicolumn{2}{c}{\textbf{Portuguese}} &
 \multicolumn{2}{c}{\textbf{Italian}} &
 \multicolumn{2}{c}{\textbf{French}} &
 \multicolumn{2}{c}{\textbf{Romanian}}\\ \cline{2-15}
Entity & NER & Pipeline & NER & Pipeline & NER & Pipeline & NER & Pipeline & NER & Pipeline & NER & Pipeline & NER & Pipeline \\ \hline 
Age & 0.910 & 0.967 & 0.944 & 0.965 & 0.971 & 0.987 & 0.963 & 0.984 & 0.969 & 0.984 & 0.933 & 0.978 & 0.840 & 0.933\\
Date & 0.973 & 0.988 & 0.999 & 0.999 & 0.965 & 0.978 & 0.989 & 0.995 & 0.985 & 0.986 & 0.991 & 0.997 & 0.915 & 0.952\\
ID & 0.930 & 0.974 & 0.974 & 0.984 & 0.978 & 0.994 & 0.978 & 0.996 & 0.980 & 0.988 & 0.966 & 0.983 & 0.893 & 0.952\\
Location & 0.803 & 0.927 & 0.797 & 0.855 & 0.870 & 0.903 & 0.958 & 0.968 & 0.971 & 0.985 & 0.868 & 0.956 & 0.596 & 0.709\\
\hline
\multicolumn{1}{l}{\textbf{Avg.}} & \textbf{0.904} & \textbf{0.964} & \textbf{0.929} & \textbf{0.951} & \textbf{0.946} &  \textbf{0.965} & \textbf{0.972} & \textbf{0.986} & \textbf{0.976} & \textbf{0.986} & \textbf{0.939} & \textbf{0.979} & \textbf{0.811} & \textbf{0.887}\\ 
\hline
PHI & 0.948 & 0.982 & 0.958 & 0.966 & 0.974 & 0.983 & 0.992 & 0.994 & 0.984 & 0.992 & 0.986 & 0.996 & 0.930 & 0.957\\
\bottomrule
\end{tabular}}
\end{table*}

\subsection{Merging Rules and Models}

A key requirement for the system was supporting fast deployments as a new de-identification project on a large unseen dataset should be measured in days, not months. This meant that models could not be tuned on newly annotated data every time, and configuration largely fell to the Chunk Merger component.

The ability to easily configure which model or rule would take priority on each entity type proves critical to overall accuracy - by composing rules (which can have high precision but low recall) with models (which may have high recall but lower precision). To measure the accuracy impact of this hybrid approach, we carried out error analyses in all seven languages quantifying the effects of contextual rules on the final metrics of the full de-identification pipeline. 

This involved comparing the de-identification with NER models versus the full de-identification pipeline (which combines both models and rules). Since not all of the PHI entities are complemented by contextual parsers, we only picked four entities: Age, Date, ID, Location. 

In some use cases, we may not need the precise entity type of any PHI in masking modes: as long as a PHI entity is detected, it doesn't matter if it's a Name or Location entity for masking. Therefore, we also evaluated the binary PHI recognition performance of our solution. All tests were run on internally curated and annotated datasets for each language which were not used in the training process. 


Results show that the full de-identification pipeline has an average improvement of around 10\% across all entities. The most drastic improvements occurred in the Location and Age entities, with improvements of 12\% and 5\% respectively. When it comes to binary PHI recognition performance, the gain was between 1 and 4\%, exceeding 95\% accuracy in all the languages supported, even exceeding 99\% in some of the languages (see Table \ref{tab:nervspp}).

Beyond the overall accuracy gain, the Chunk Merger has proven useful in tackling several longstanding challenges in medical text de-identification. Consider this text:\\

\textit{John was Diagnosed with Parkinson's by Dr. Hopkins at John Hopkins Hospital.}\\

There are four entities here: the patient's name, their diagnosis (Parkinson's disease), the doctor's name, and the hospital. The diagnosis is not PHI and it's important to keep it in the de-identified text, otherwise the utility of the whole note diminishes. The three other entities also have to be corrected, classified and obfuscated to separate names.

This is addressed in the proposed system by applying multiple models and rules, and then using the chunk merger to prioritize conflict resolution. To resolve the particular challenge in the above example, we leverage a pre-trained clinical NER model that detects disease names in the pipeline, and use the Chunk Merger to override overlapping identified entities by giving the disease model a higher priority than the PHI detection model. This way, the disease name will have been identified both as a disease and a patient name, but the patient name label will have been discarded in the merging process.

\section{Conclusion}
\label{sec:conc}

De-identification of electronic health records (EHR) is essential to enabling secondary use of medical data for a broad range of safety, research, and public health use cases. Recent solutions achieve human-level accuracy for de-identifying free-text clinical notes on research datasets, but gaps remain in reproducing this in large-scale, real-world settings. 


This study describes a medical text de-identification system that is first to achieve the trifacta of state-of-the-art accuracy of the data science models, fulfilling the engineering requirements of high-compliance production systems, and independently certified success in multiple real-world deployments. Delivering fully automated de-identification in practice requires solving challenges beyond accuracy - consistent obfuscation, fast deployments, scalability on commodity hardware, configurability, and the ability to support a new language within a few weeks.

While further work remains to apply the solution more broadly, quickly, and cheaply, this and similar solutions are already being widely deployed, fundamentally changing the landscape of clinical data availability. This unlocks new opportunities for the secondary use of medical data in a broad range of safety, research, and public health use cases, in a safer and compliant manner.

\clearpage


\bibliography{jmlr-sample}

\clearpage
\appendix

\section{Performance on 2014 i2b2 Challenge, 13-labels granular version dataset}\label{apd:first} 
Table \ref{tab:i2b2_stats_subentity} illustrate performance metrics on granular entities for the 2014 i2b2 challenge.

\begin{table}[hbt!]
\caption{NER metrics on 2014 i2b2 Challenge using more granular 13 labels.}
\label{tab:i2b2_stats_subentity}
\begin{tabular}{lccc}
\toprule
Entity & Precision & Recall & F1\\
\midrule
Patient & 0.958 & 0.977 & 0.967\\
Hospital & 0.98 & 0.965 & 0.972\\
Date & 0.996 & 0.995 & 0.996\\
Organization & 0.927 & 0.826 & 0.874\\
City & 0.953 & 0.944 & 0.949\\
Street & 0.998 & 0.995 & 0.996\\
Username & 0.989 & 0.921 & 0.954\\
Device & 0.5 & 0.2 & 0.286\\
Fax & 1.0 & 0.667 & 0.8\\
Idnum & 0.873 & 0.948 & 0.909\\
State & 0.949 & 0.99 & 0.969\\
Email & 0.0 & 0.0 & 0.0\\
Zip & 0.979 & 0.986 & 0.982\\
Medicalrecord & 0.989 & 0.971 & 0.98\\
Other & 0.867 & 0.619 & 0.722\\
Profession & 0.968 & 0.887 & 0.925\\
Phone & 0.983 & 0.974 & 0.978\\
Country & 0.896 & 0.945 & 0.92\\
Healthplan & 1.0 & 1.0 & 1.0\\
Doctor & 0.986 & 0.974 & 0.98\\
Age & 0.97 & 0.959 & 0.964\\
\midrule
\textbf{Macro-Avg.} & & & \textbf{0.863}\\
\textbf{Micro-Avg.} & & & \textbf{0.978}\\
\bottomrule
\end{tabular}
\end{table}

\section{Metrics on multiple languages}
\label{apd:second}
Table \ref{tab:stats_multi_lang} shows performance metrics on seven different languages.

\begin{table*}[!bp]
\caption{Metrics on multiple languages.}
\scalebox{1.0}{
\label{tab:stats_multi_lang}
\begin{tabular}{lccccccc}
\toprule
Entity & English & German & French & Spanish & Italian & Portuguese & Romanian\\
\midrule
Patient & 0.9 & 0.97 & 0.94 & 0.92 & 0.91 & 0.95 & 0.87\\
Doctor & 0.94 & 0.98 & 0.99 & 0.92 & 0.92 & 0.93 & 0.96\\
Hospital & 0.91 & 1.00 & 0.94 & 0.86 & 0.90 & 0.90 & 0.8\\
Date & 0.98 & 1.00 & 0.98 & 0.99 & 0.98 & 0.98 & 0.91\\
Age & 0.94 & 0.99 & 0.86 & 0.98 & 0.98 & 0.98 & 0.97\\
Profession & 0.84 & 1.00 & 0.81 & 0.91 & 0.89 & 0.90 & 0.83\\
Organization & 0.77 & 0.94 & 0.77 & 0.83 & 0.74 & 0.97 & 0.37\\
Street & 0.98 & 0.98 & 0.90 & 0.94 & 0.98 & 1.00 & 0.99\\
City & 0.83 & 0.99 & 0.86 & 0.84 & 0.97 & 0.98 & 0.96\\
Country & 0.81 & 0.98 & 0.90 & 0.87 & 0.93 & 0.91 & 0.82\\
Phone & 0.94 & 0.88 & 0.98 & 0.90 & 0.98 & 0.99 & 0.98\\
Username & 0.92 & 1.00 & 0.92 & 0.74 & 0.91 & 0.88 & -\\
ZIP & 0.99 & - & 1.00 & 0.99 & 0.99 & 0.99 & 0.98\\
\midrule
\textbf{Macro-Avg.} & \textbf{0.904} & \textbf{0.901} & \textbf{0.912} & \textbf{0.899} & \textbf{0.929} & \textbf{0.951} & \textbf{0.803}\\
\bottomrule
\end{tabular}}
\end{table*}

\section{Comparison with ChatGPT}\label{apd:third}

\subsection{Study Design}
\bigbreak
\textbf{Data Selection:} We selected 25 clinical discharge notes from the 2014 i2b2 Deid Challenge dataset and annotated further to reduce label errors.
\bigbreak
\textbf{Scope:} The following entities are considered during this comparison: ‘ID’, ‘DATE’, ‘AGE’, ‘PHONE’, ‘PERSON’, ‘LOCATION’, ‘ORGANIZATION’.
\bigbreak
\textbf{Model Preparation:} To prevent data leakage, we examined the dataset of the selected NER models for PHI detection, retraining some models from scratch for evaluation purposes.
\bigbreak
\textbf{Prediction Collection:} We ran the prompts per entity in a few shot settings, and collected the predictions from ChatGPT, which were then compared with the ground truth annotations.
\bigbreak
\textbf{Performance Comparison:} We obtained predictions from the corresponding NER models and Contextual Parsers and compared those with the ground truth annotations as well (even a single token overlapping was considered a hit).
\bigbreak
\textbf{Reproducibility:} All prompts, scripts to query ChatGPT API (ChatGPT-3.5-turbo, temperature=0.7), evaluation logic, and results will be shared publicly at a Github repo after blind review process. Additionally, a detailed Colab notebook to run De-Identification modules step by step will also be shared publicly.
\bigbreak
\subsection{Results}

\begin{figure*}[h]
  \centering
  \includegraphics[width=\linewidth]{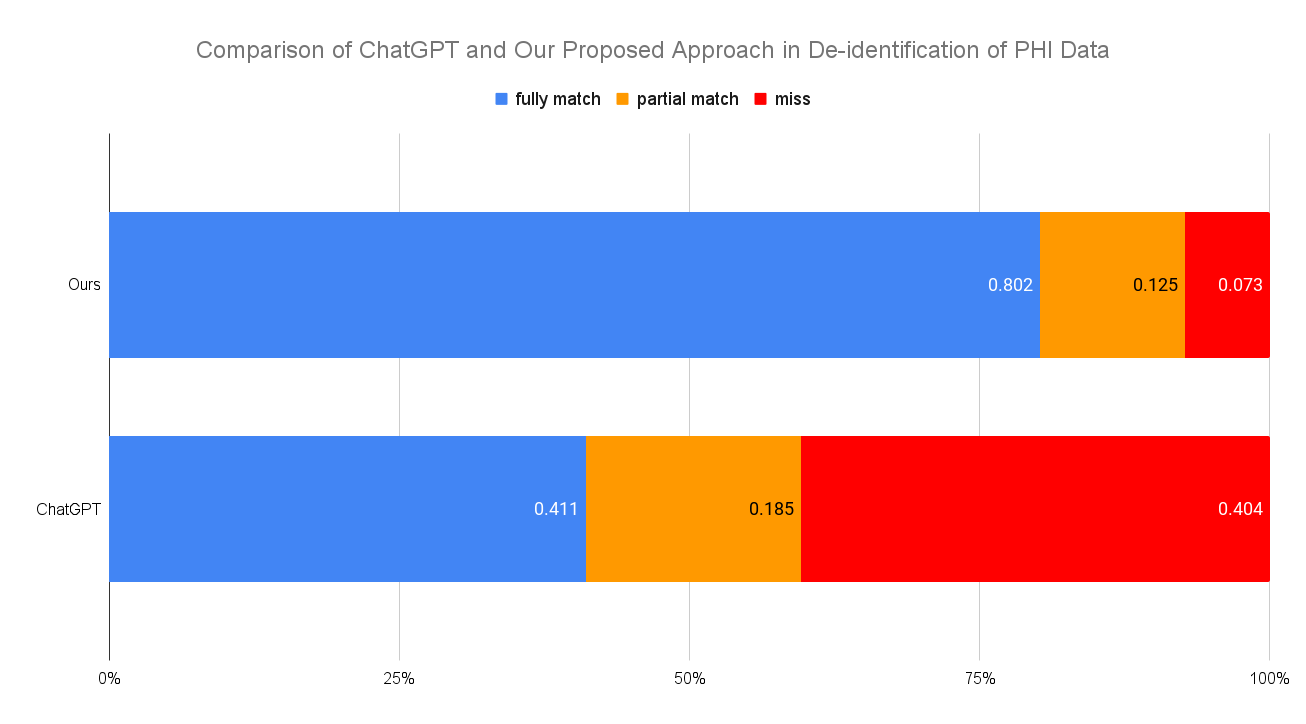}
  \caption{The performance of ChatGPT and our proposed approach in de-identifying PHI data within clinical discharge notes. The comparison includes the total number of entities, accuracy rates, and the percentage of partially and fully matched entities for both tools. Our approach demonstrates superior performance with a 93\% accuracy rate compared to ChatGPT’s 60\% accuracy.}
  \label{fig:chatgpt}
\end{figure*}

As can be seen in Figure \ref{fig:chatgpt}, out of 562 sensitive entities, ChatGPT failed to identify 227, resulting in an accuracy of approximately 60\%. Among its findings, 20\% were partially matched (with at least one token overlapping with the ground truth), while 41\% were fully matched. We then processed the same documents using our proposed De-Identification pipeline. Out of the 562 sensitive entities, it failed to find 41, achieving an accuracy of around 93\%. Of these findings, 14\% were partially matched (with at least one token overlapping with the ground truth), and 79\% were fully matched.

\end{document}